\documentclass[letterpaper, 10 pt, conference]{ieeeconf}  

\IEEEoverridecommandlockouts                              

\usepackage{graphics} 
\usepackage{epsfig} 
\usepackage{mathptmx} 
\usepackage{times} 
\usepackage{amsmath} 
\usepackage{amssymb}  
\usepackage{cite} 
\usepackage{multirow}
\usepackage{xcolor}
\usepackage{booktabs}
\usepackage{tabularx}
\usepackage{adjustbox}
\usepackage{tikz}
\title{\LARGE \bf
Three Degree-of-Freedom Soft Continuum Kinesthetic Haptic Display for Telemanipulation Via Sensory Substitution at the Finger 
}

\author{Jiaji Su$^{1}$, Kaiwen Zuo$^{2}$ and Zonghe Chua$^{1}$
\thanks{$^{1}$J.\,Su and Z.\,Chua are with the Department of Electrical, Computer, and Systems Engineering, Case Western Reserve University, Cleveland, OH 44106, USA
        {\tt\small jxs1778@case.edu ,zxc703@case.edu}}%
\thanks{$^{2}$K.\,Zuo is with the Department of Mechanical and Aerospace Engineering, Case Western Reserve University, Cleveland, OH 44106, USA
        {\tt\small kxz365@case.edu}}%
}

\newcommand{\hlight}[1]{\textcolor{black}{#1}}
\newcommand{\changed}[1]{\textcolor{black}{#1}}
\newcommand{\revised}[1]{\textcolor{black}{#1}}

\newcommand\copyrighttext{%
  \footnotesize \textcopyright 2024 IEEE. Personal use of this material is permitted.
  Permission from IEEE must be obtained for all other uses, in any current or future 
  media, including reprinting/republishing this material for advertising or promotional 
  purposes, creating new collective works, for resale or redistribution to servers or 
  lists, or reuse of any copyrighted component of this work in other works.}
  
\newcommand\copyrightnotice{%
\begin{tikzpicture}[remember picture,overlay]
\node[anchor=south,yshift=10pt] at (current page.south) {\fbox{\parbox{\dimexpr\textwidth-\fboxsep-\fboxrule\relax}{\copyrighttext}}};
\end{tikzpicture}%
}

\begin{document}
\bstctlcite{IEEEexample:BSTcontrol} 

\maketitle
\thispagestyle{empty}
\pagestyle{empty}
\copyrightnotice

\begin{abstract}

Sensory substitution is an effective approach for displaying stable haptic feedback to a teleoperator under time delay. The finger is highly articulated, and can sense movement and force in many directions, making it a promising location for sensory substitution based on kinesthetic feedback. However, existing finger kinesthetic devices either provide only one-degree-of-freedom feedback, are bulky, or have low force output. Soft pneumatic actuators have high power density, making them suitable for realizing high force kinesthetic feedback in a compact form factor. We present a soft pneumatic handheld kinesthetic feedback device for the index finger that is controlled using a constant curvature kinematic model. \changed{It has respective position and force ranges of $\pm$3.18\,mm and $\pm$1.00\,N laterally, and $\pm$4.89\,mm and $\pm$6.01\,N vertically, indicating its high power density and compactness. The average open-loop radial position and force accuracy of the kinematic model are 0.72\,mm and 0.34\,N.} Its 3\,Hz bandwidth makes it suitable for moderate speed haptic interactions in soft environments. We demonstrate the three-dimensional kinesthetic force feedback capability of our device for sensory substitution at the index figure in a virtual telemanipulation scenario. 

\end{abstract}

\section{INTRODUCTION}
Haptic feedback plays a valuable role in telerobotic manipulation, enabling an operator to significantly reduce their applied forces, thereby improving task performance and increasing safety \cite{wagner_benefit_2007}. In remote telemanipulation applications, delay can make delivering stable, and transparent haptic feedback to the hand challenging  \cite{hashtrudi-zaad_analysis_2001}. One effective solution is sensory substitution, where a device presents kinesthetic force feedback in a different modality such as vision \cite{kitagawa_effect_2005}, normal indentation \cite{sarac_augmented_2021}, or squeezing \cite{pezent_tasbi_2019,machaca_towards_2022}. Researchers have also applied sensory substitution to the fingertips to leverage their high tactile sensitivity \cite{quek_sensory_2015,schorr_fingertip_2017}. 

More broadly, the finger as a whole provides multi-directional kinesthetic and positional sensing, making it a promising location for delivering sensory substitution \cite{tan_manual_1995,barbagli_haptic_2006}. However, providing kinesthetic sensory substitution feedback to the whole finger has been less explored. Most wearable finger haptic devices lack multi-directional force rendering capabilities as such configurations have traditionally been very bulky \cite{Koyama2002,Frisoli2006,iqbal2015four}. Designers instead opt to keep transmission systems compact to improve wearability, and provide single degree-of-freedom grasp forces to each finger \cite{wang_survey_2023}. Recently, Shabani et al. \cite{shabani_human-centered_2022} proposed a compact tendon-driven wearable finger haptic device with multi-directional force rendering capabilities constrained along the tendon routings. 

Handheld devices offer another hand-grounded form factor with finger kinesthetic feedback capability \cite{gonzalez-franco_taxonomy_2022}. Like many wearable devices, these mainly focused on providing one-degree-of-freedom (DoF) feedback \cite{sun_pacapa_2019,sinclair_capstancrunch_2019,choi_claw_2018,gonzalez_x-rings_2021}. A notable exception has been the Foldaway origami device from Mintchev et al. \cite{mintchev_portable_2019}, which can provide three-DoF force feedback to a single digit. Foldaway's reliance on DC motors, which have a relatively low power density \cite{mavroidis2002development}, results in a lower overall force range, and precludes further miniaturization of the drive system into form factors seen in applications such as telesurgery \cite{innocenti2023first,freschi2013technical}, where haptic interactions can potentially involve multiple digits, and interface inertia can influence performance \cite{nisky_effects_2014}.
\begin{figure}[t]
    \vspace{1.7mm}
    \centering
    \includegraphics[width=0.9\linewidth]{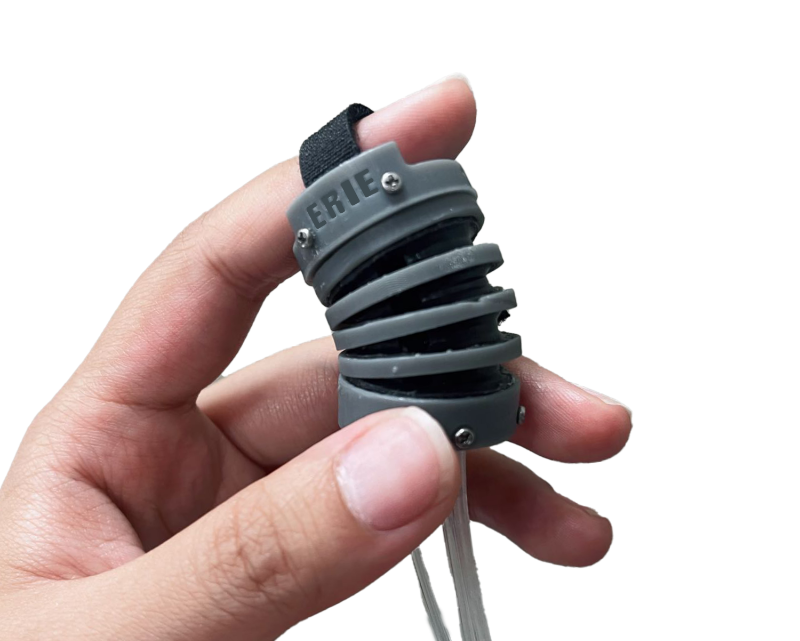}
    \caption{The fabricated soft pneumatic continuum haptic device used in a handheld configuration for sensory substitution to the finger.}
    \vspace{-2em}
    \label{device}
\end{figure}

Soft pneumatic actuators (SPAs) have higher power density than DC motors, particularly if the pneumatic source is off-board \cite{mavroidis2002development}. Thus, they present a promising approach to increasing the force output of the multi-DoF finger kinesthetic haptic device. However, researchers have only mainly explored compact fingertip tactile feedback devices \cite{hachisu_vacuumtouch_2014,hashem_soft_2022,zhakypov_fingerprint_2022}, and wearable single-DoF kinesthetic devices \cite{polygerinos2015soft,jadhav2017soft,lim2023bidirectional}. In contrast, soft continuum robots can leverage multiple pneumatic actuators to realize multi-DoF movement \cite{falkenhahn_dynamic_2015,chauhan_origami-based_2021,caasenbrood_control-oriented_2023}. Thus, similar to how traditional kinesthetic devices are variants of serial link or parallel link rigid manipulators, we can employ soft pneumatic robotic systems toward kinesthetic haptic device applications. 

This paper introduces a handheld soft continuum haptic device that provides three-DoF kinesthetic force feedback to the index finger (Fig.\,\ref{device}). \revised{It consists of three parallel SPAs, each containing foldable air chambers with miniature 3D-printed structures. Compared to existing designs, it has notably high power density, enabling high force output in a portable form factor.} We present and evaluate the constant curvature kinematic model \cite{webster_design_2010} for Jacobian-based open-loop position control of the fingertip interaction point and demonstrate the feasibility of our device for haptic sensory substitution to the index finger in a virtual telemanipulation scenario.

\section{Hardware Design and Fabrication}

\subsection{Design of the Soft Continuum Haptic Device}

\begin{figure}
    \vspace{1.7mm}
    \centering
    \includegraphics[width=\linewidth]{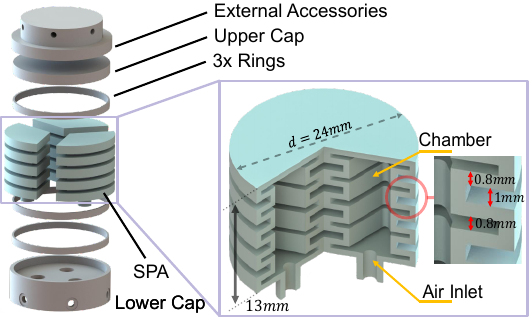}
    \caption{CAD model of the soft continuum haptic device. Inset depicts the assembled configuration of three SPA units.}
    \label{fig:device_design}
\end{figure}

To achieve three-DoF force feedback using SPAs, we use a continuum robotics approach, where three SPAs are arranged in parallel to drive a single end plate in extension and bending \cite{caasenbrood_control-oriented_2023}. Electronic adjustment of the internal pressure within the SPAs then controls their lengths. To achieve the rapid response and miniaturization required for a centimeter-scale handheld finger haptic device, we utilize 3D printing with elastomeric materials to realize millimeter-scale internal features and wall thicknesses. The design of individual SPAs and the overall device is summarized in Fig.\,\ref{fig:device_design}.

Each SPA actuator consists of three simultaneously inflatable cells. The cross-section of the inflatable cells spans a 120$^\circ$ sector of a circle with 12\,mm radius. The wall thickness of the cells is 0.8\,mm. When the chambers are depressurized to atmosphere, the SPA actuator adopts a folded state with a height of 13\,mm, and a 1\,mm gap between the chambers which reduces self-interference during bending. The SPAs are arranged in parallel to form a single device whose cross-section is a full circle. Three rigid rings, one for each outward-facing fold of the inflatable cells, with wall thickness of 0.5\,mm are added to the circumference of the device to constrain radial deformation during pressurization. These rings also reinforce against fatiguing of the folds as the device expands axially. Rigid upper and lower caps are designed to limit unwanted deformation at each end and allow the connection of external accessories and pneumatic lines.

\subsection{Fabrication}


The SPAs are printed using Formlabs Silicone 40A resin on a Form 3 printer (Formlabs, Somerville, MA). During the manufacturing process, the SPAs are horizontally oriented with their arced surface facing toward the print bed. This enables the support structure points to be placed only on the outer walls, thus avoiding the need for internal support that cannot be removed. The printed parts are washed in an isopropyl alcohol (IPA) bath for 30 minutes. The inner cavities are then flushed with IPA before they are UV-cured. Lastly, the parts are placed in a water-filled glass beaker at 60°C for 20 minutes to achieve the required stiffness. The three actuators are bonded together using cyanoacrylate (Bob-Smith Industries, Eustis, FL). The rigid ring reinforcements were made of Formlabs Grey V4.

\subsection{Pneumatic Control System}

\begin{figure}
    \centering
    \includegraphics[width=0.9\linewidth]{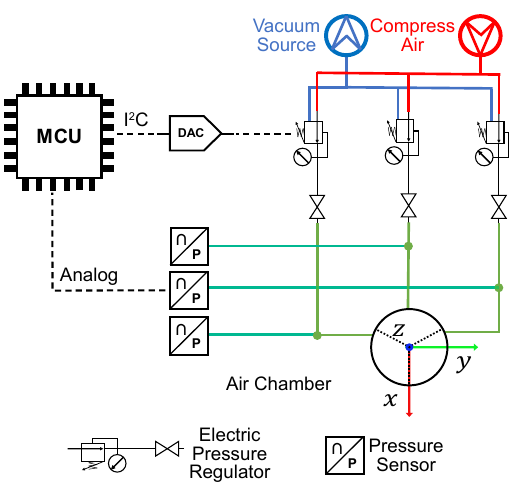}
    \caption{Schematic of the pneumatic actuation and sensing system}
    \label{fig:mechatronics}
\end{figure}

The pressure in each SPA is regulated by an ITV0030 pressure regulator (SMC, Chiyoda City, Tokyo, Japan). An air compressor (Delwalt, Towson, MD) is used to provide a 500\,kPa positive pressure source. Furthermore, a vacuum chamber connected to a vacuum pump (BVV, Naperville, IL)  provides -95\,kPa to the exhaust port of the regulator to ensure the desired air flow rate. An Arduino Mega2560 provides a control input to the regulator via an MCP4728 digital-to-analog converter (Microchip Tech. Inc., Chandler, AZ). The output pressure is monitored by an ADP5200 analog pressure sensor  \revised{(Panasonic, Kadoma, Osaka, Japan)} for diagnostics. The schematic of the control system, and the device local coordinate frame with respect to the actuator arrangement is shown in Fig.\,\ref{fig:mechatronics}. \changed{The local frame of the device is defined such that the origin is located at the base of the device with the z-axis aligned along the main axis of the device. The x-axis points outward along the boundary of two SPAs.
} 

\section{Kinematic Modeling and Control}
 
\subsection{Constant Curvature Model} 
\begin{figure}[t]
  \vspace{1.7mm}
  \centering
    \includegraphics[width=\linewidth]{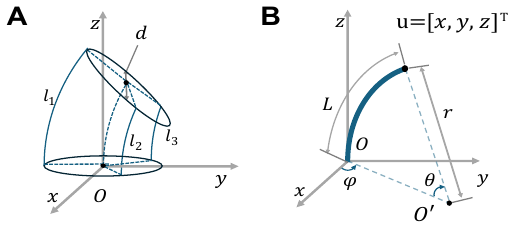}
  \caption{Constant curvature model (A) configuration and (B) geometric space descriptions.}
  \label{fig:models}
\end{figure}

In the forward kinematic model, we assume that the device is a single-segment continuum robot and apply the piecewise constant curvature model \cite{webster_design_2010}. The SPA pressures are related to their length by a mapping function
\begin{equation}
\label{eqn:p_to_l_map}
l_i = f(p_i) \text{,}
\end{equation}
where $p_i$ and $l_i$ are the pressure and the length of the i\textsuperscript{th} SPA, respectively. This function can be empirically determined by characterizing the pressure-to-length response of the individual SPAs to account for manufacturing variations. When approximated by a polynomial or exponential, this mapping can be inverted to obtain the length-to-pressure mapping. The configuration space of the continuum haptic device is described by the arc parameters $\kappa$, $\varphi$, and $\theta$, which are the radius of curvature, the rotation about the z-axis, and the sector angle subtended by the arc, respectively (See Fig.\,\ref{fig:models}). The relationship between chamber lengths and arc parameters are 
\begin{equation}
\kappa (q) = \frac{{2\sqrt {l_1^2 + l_2^2 + l_3^2 - {l_1}{l_2} - {l_1}{l_3} - {l_2}{l_3}} }}{{d({l_1} + {l_2} + {l_3})}} \text{,}
\end{equation}
\begin{equation}
\varphi (q) = {\tan ^{ - 1}}\Bigl(\frac{{\sqrt 3 ({l_2} + {l_3} - 2{l_1})}}{{3({l_2} - {l_3})}}\Bigl) \text{, and}
\end{equation}
\begin{equation}
    \theta (q) = \frac{{2\sqrt {l_1^2 + l_2^2 + l_3^2 - {l_1}{l_2} - {l_1}{l_3} - {l_2}{l_3}} }}{{3d}} \text{ ,}
\end{equation}
with the backbone length $L$ shown in Fig.\,\ref{fig:models}A and B, being the average of the the SPA lengths
\begin{equation}
L=\frac{{{l_1} + {l_2} + {l_3}}}{3} \text{ .} 
\end{equation}
The homogenous transformation describing the tip pose is then
\begin{equation}
\begin{array}{c}
T = \left[ \begin{array}{cc}
{R_z(\varphi)} & 0 \\
0 & 1 \\
\end{array} \right]
\left[ \begin{array}{cc}
{R_y(\theta)} & p \\
0 & 1 \\
\end{array} \right]
\end{array} \text{, where}
\end{equation}
\begin{equation*}
    \ p = \left[ r(1 - \cos \theta) \quad 0 \quad r \sin \theta \right]^T \text{ and }
\end{equation*}
\begin{equation*}
     \kappa = 1/r \text{,} 
\end{equation*}
which results in the following position description of the control point 
\begin{equation}
u =
\begin{bmatrix}
\frac{\cos \phi (1 - \cos \theta)}{\kappa} \\
\frac{\sin \phi (1 - \cos \theta)}{\kappa} \\
\frac{\sin \theta}{\kappa}
\end{bmatrix} \text{.}
\end{equation}

\subsection{Jacobian-based Control Approach}

To perform position control, we use the above kinematic model to perform Jacobian-based resolved-rate motion control \cite{whitney1969resolved} for the soft haptic device, using the equation
\begin{equation}
\label{eqn:jacobian}
\dot{l_n} = J(l_n)^{+} \dot{u}_n
\end{equation}
where $l_n = \begin{bmatrix}l_{1,n} & l_{2,n} & l_{3,n}\end{bmatrix}^T$ is the vector of SPA lengths at the n\textsuperscript{th} timestep, $J^+$ is the Moore-Penrose pseudoinverse of the Jacobian. The velocity $\dot{u}_n$ is obtained by computing the direction unit vector between the current control point position and the commanded desired position in task space
\begin{equation}
\dot{u}_n = \frac{(u_{n,\text{d}}-u_{n})}{|u_{n,\text{d}}-u_{n}|} \text{ ,}
\end{equation}
where $u_{n,d}$ is the desired control point position. At each time step, the Jacobian is computed using backward numerical differentiation. The length of the SPA at $n+1$ is then 
\begin{equation}
l_{n+1} = l_n + c\,\dot{l}_n
\end{equation}
where $c$ is a scaling constant. By adjusting the value of this constant, the speed at which the control point moves in the task space can be controlled. Based on empirical testing for tracking performance, the value of $c$ was set to 10. The resultant length vector is used to set the required length of the SPAs by inverting the fitted mapping function defined in (\ref{eqn:p_to_l_map}). To avoid the singularity when the device is in an unbent configuration and the radius of curvature $\kappa$ is zero, we add a small offset of 0.0001\,mm.

\section{Experimental System Validation}

\label{sec:exp_validation}
\begin{figure}
    \vspace{1.7mm}
    \centering
    \includegraphics[width=\linewidth]{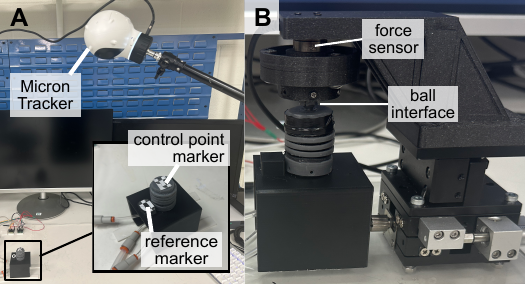}
    \caption{Experimental setup for (A) position, inset shows the close-up of the reference and control point visual markers, and (B) force output validation.}
    \label{fig:experimentalsetup}
\end{figure}

\subsection{Path Accuracy and Dynamic Response}
\label{sec:pos_validation}
The experimental setup for measuring the position of the device control point and dynamic response is shown in Fig.\,\ref{fig:experimentalsetup}A. The prototype of the soft haptic device is mounted on a 3D-printed rigid frame, and a stereo camera, MicronTracker (ClaroNav, Toronto, ON, Canada), is used to track the position of the device control point. We attached a marker to the flat-top cap of the device, with the origin of the marker coinciding with the control point corresponding to the center of the cap. Another marker was attached to the rigid frame to establish a reference coordinate system. During the experiment, the position of device control point with respect to the reference coordinate system was recorded at a frequency of 56\,Hz. 

The open-loop position tracking accuracy of our device was characterized by commanding it to follow pre-specified paths within the feasible work range of the device. For this, we selected three circles in the workspace, with respect to a starting position in which the device is pre-pressurized to 25\,kPa for an initial vertical device height of 23\,mm. These consisted of an upper circle 2.5\,mm above the starting plane with a radius of 3\,mm, a middle circle on the starting plane with a radius of 5\,mm, and a lower circle 2.5\,mm below the plane with a radius of 3\,mm. For the upper and lower small circles, we selected six points at 60° intervals, and for the middle large circle, we selected twelve points at 30° intervals. The path error was computed by finding the perpendicular distance of the sampled control point positions to the line vector in the commanded direction. The range of the device in each direction is computed as a radius from the center of a sphere corresponding to the device's initial configuration.

To characterize the dynamic response of the system, we conducted a bandwidth test. The device frequency responses were collected at 0.1\,Hz, 0.5\,Hz, and from 1\,Hz to 4\,Hz in 0.5\,Hz increments, and from 4\,Hz to 10\,Hz in 1\,Hz increments. 3 \,mm amplitude sinusoidal oscillations were commanded in the x-, y-, and z-directions. The measured amplitudes were estimated by fitting the parameters $A$,$f$, $\phi$, and $x_0$ of a sinusoidal function
\begin{equation}
x_{t} = A  \cdot \sin(2 \pi f t + \phi) + x_{0}
\label{Eq: sine_fit}
\end{equation}
using \texttt{scipy.optimize.curve\_fit}. The magnitude ration was computed as
\begin{equation}
    \text{MR} = \frac{A}{A_c} \text{ ,}
\end{equation}
 where $A_{\text{max}}$ was the amplitude of the fitted function, and $A_c$ was the commanded amplitude.

\subsection{Force Output and Dynamic Range}

As depicted in Fig.\,\ref{fig:experimentalsetup}B, the block force of the device was measured using a six-axis force sensor, ATI Nano17 (ATI Industrial Automation, Apex, NC). The sensor was fixed on a 3-axis linear stage (Shenzhen Tianzhongtian Trading Co., Ltd., Guangdong, China) using 3D-printed components. The soft haptic device was mounted on the same frame. The device tip was aligned coaxially with the sensor and connected through a 3D-printed ball socket joint. Due to the ball socket mechanism's inability to transmit torque, the force output and torque of the device tip were decoupled, allowing the sensor to measure only the force output of the device. 

We conducted a force output characterization experiment by using the position commands to push the device control point in the same target directions as used in the path accuracy experiment. The device's movement was commanded until any of the SPA's internal pressure reached their safety limit of 50kPA. Three measurements were taken per commanded direction with the range and positional error computed in the same way described in Sect.\,\ref{sec:pos_validation}. Likewise, a force output bandwidth test was conducted as per the specifications described in Sect.\,\ref{sec:pos_validation}.

\begin{figure*}[t]
    \vspace{1.7mm}
    \centering
    \includegraphics[width=\linewidth]{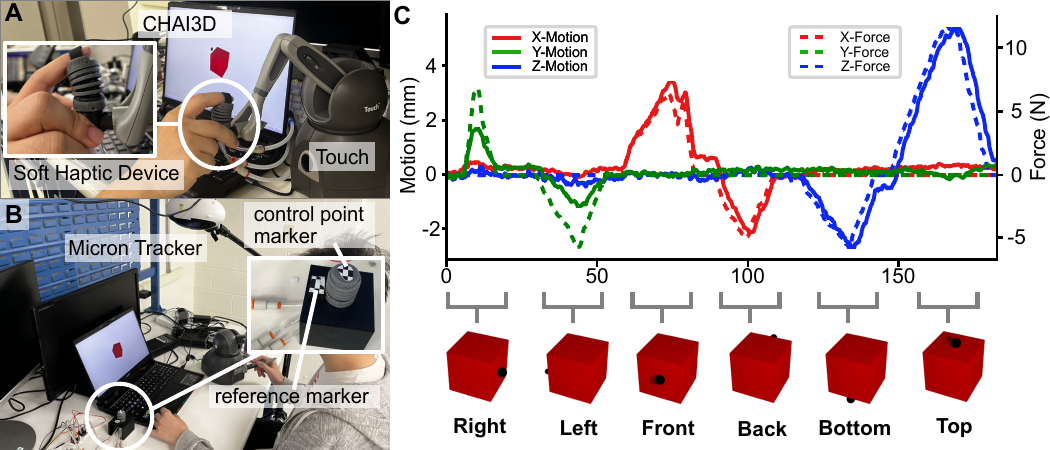}
    \caption{(A) Setup for virtual reality interaction using CHAI3D framework and the soft haptic device mounted on Touch. (B) Free motion tracking during interaction in CHAI3D (C) Interaction forces generated in CHAI3D and resultant position outputs from the device during different directional interactions with a virtual cube.}
    \label{fig:demo}
\end{figure*}

\newcolumntype{Y}{>{\centering\arraybackslash}p{1.57cm}}
\newcolumntype{A}{>{\centering\arraybackslash}p{0.6cm}}
\newcolumntype{B}{>{\centering\arraybackslash}p{1cm}}
\newcolumntype{L}{>{\centering\arraybackslash}X}
\setlength{\tabcolsep}{1.2pt}
\begin{table*}[t]
    \centering
    \caption{Force and positional range and average path error across different levels in different directions corresponding to those shown in Fig.\,\ref{fig:pos_force_output}. Ranges are reported as a radius from the center of a sphere.}
    \begin{adjustbox}{max width=\textwidth}
    \begin{tabularx}{\textwidth}{ABBBBBBYYYYYY}
        \toprule
        \multirow{4}{*}{\centering\textbf{Path}} & \multicolumn{3}{c}{\textbf{Positional Radial Range (mm)}} & \multicolumn{3}{c}{\textbf{Force Radial Range(N)}} & \multicolumn{3}{c}{\textbf{Positional Path Error (mm)}} &\multicolumn{3}{c}{\textbf{Force Path Error (mm)}} \\
        \cmidrule(lr){2-4}\cmidrule(lr){5-7}\cmidrule(lr){8-10}\cmidrule(lr){11-13}
        & \textbf{Lower Range} & \textbf{Middle Range} & \textbf{Upper Range} & \textbf{Lower Range} & \textbf{Middle Range} & \textbf{Upper Range} & \textbf{Lower Error} & \textbf{Middle Error} & \textbf{Upper Error} & \textbf{Lower Error} & \textbf{Middle Error} & \textbf{Upper Error} \\
        \midrule
        0 & -- & 5.60 & -- & -- & 1.13 & -- & -- & 0.47 ± 0.25 & -- & -- & 0.07 ± 0.03 & -- \\ 
        30 & 3.31 & 6.00 & 4.98 & 2.94 & 1.01 & 2.31 & 0.59 ± 0.38 & 0.73 ± 0.37 & 0.23 ± 0.12 & 0.04 ± 0.02 & 0.10 ± 0.05 & 0.14 ± 0.07 \\
        60 & -- & 5.97 & -- & -- & 1.06 & -- & -- & 0.92 ± 0.66 & -- & -- & 0.24 ± 0.14 & -- \\
        90 & 3.97 & 4.68 & 4.59 & 2.90 & 1.32 & 2.23 & 0.85 ± 0.51 & 0.73 ± 0.53 & 0.55 ± 0.36 & 0.15 ± 0.05 & 0.09 ± 0.05 & 0.24 ± 0.13 \\
        120 & -- & 4.48 & -- & -- & 1.29 & -- & -- & 0.44 ± 0.29 & -- & -- & 0.05 ± 0.03 & -- \\
        150 & 3.76 & 5.08 & 3.96 & 1.94 & 1.48 & 2.23 & 0.66 ± 0.43 & 0.47 ± 0.25 & 0.47 ± 0.35 & 0.27 ± 0.13 & 0.06 ± 0.04 & 0.48 ± 0.31 \\
        180 & -- & 4.88 & -- & -- & 1.36 & -- & -- & 0.67 ± 0.26 & -- & -- & 0.30 ± 0.14 & -- \\
        210 & 3.18 & 4.84 & 3.88 & 2.91 & 1.31 & 2.45 & 0.63 ± 0.50 & 0.54 ± 0.28 & 0.31 ± 0.16 & 0.06 ± 0.03 & 0.14 ± 0.07 & 0.28 ± 0.14 \\
        240 & -- & 5.68 & -- & -- & 1.20 & -- & -- & 0.46 ± 0.23 & -- & -- & 0.10 ± 0.06 & -- \\ 
        270 & 3.99 & 5.89 & 4.76 & 2.78 & 1.09 & 2.26 & 0.87 ± 0.50 & 0.55 ± 0.27 & 0.41 ± 0.21 & 0.26 ± 0.16 & 0.13 ± 0.04 & 0.42 ± 0.20 \\ 
        300 & -- & 5.75 & -- & -- & 1.17 & -- & -- & 1.03 ± 0.55 & -- & -- & 0.17 ± 0.08 & -- \\ 
        330 & 4.20 & 5.32 & 4.83 & 2.79 & 1.04 & 2.77 & 0.86 ± 0.52 & 0.82 ± 0.47 & 0.35 ± 0.16 & 0.10 ± 0.03 & 0.19 ± 0.06 & 0.49 ± 0.30 \\ 
        Z$+$ & -- & 9.25 & -- & -- & 6.79 & -- & -- & 1.61 ± 0.26 & -- & -- & 0.13 ± 0.05 & -- \\ 
        Z$-$ & -- & 4.89 & -- & -- & 6.01 & -- & -- & 1.02 ± 0.65 & -- & -- & 0.27 ± 0.19 & -- \\ 

        \bottomrule
    \end{tabularx}
    \end{adjustbox}
    \label{tab:combined_ring}
\end{table*}

\subsection{Virtual Reality Interaction}

 The CHAI3D framework was used to render a virtual cube for haptic interaction. The soft haptic device was mounted on a Touch Haptic Device (3D Systems Inc., Rockhill, SC), which was used only for its position input capability. The device is a traditional serial link position input device, but its small manipulandum can function as a proxy for a handheld form factor. When the cursor contacts the virtual cube from different directions, the force generated in CHAI3D is scaled into a commanded displacement. This displacement is converted into pressures using resolved-rate motion control and commanded to the device to deliver sensory substitution position feedback to the index finger (Fig.\,\ref{fig:demo}A). The mapping from CHAI3D force to the commanded device position was 4.75\,N:1\,mm. To showcase the positional variation of the device's control point in response to the virtual interaction forces, we measure the device displacement in free motion during a CHAI3D interaction as shown in Fig.\,\ref{fig:demo}B. 

\section{RESULTS AND DISCUSSION}

\begin{figure}
    \vspace{1.7mm}
    \centering
    \includegraphics[width=\linewidth]{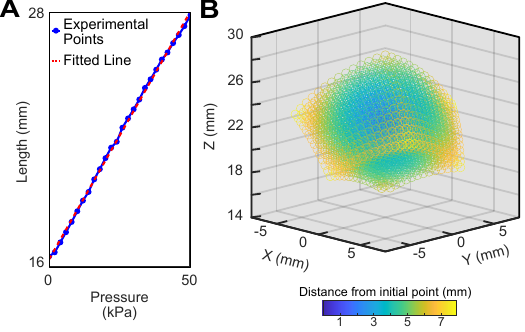}
    \caption{(A) Plot of the experimental length vs. pressure data of the device when all SPAs are equally pressurized. (B) Work space simulation based on forward kinematic model. The initial point is defined as the point where the device is pre-pressurized to a height of 23\,mm.}
    \label{fig:workspace}
\end{figure}

\revised{According to the pressure-to-length characterization experiment in which all SPAs were equally pressurized and the device's vertical extension measured, the pressure to length response of the SPAs exhibit good linearity. The linearity observed in Fig. \ref{fig:workspace}A is attributed to the air chamber's structural design, which allows the device to extend and contract mainly by folding and unfolding its walls instead of relying solely on the material's tensile properties.} For the 26 data points ranging from 0 to 50\,kPa, a linear fit was performed, resulting in $l_i = 0.23 p_i + 16.35$ (Eqn.\,\ref{eqn:p_to_l_map}). Based on the constant curvature model, we sweep all pressure combinations within each chamber from 0 to 50\,kPa in MATLAB, yielding the workspace shown in Fig.\,\ref{fig:workspace}B. In the workspace simulation, the preload pressure was 25\,kPa in each SPA corresponding to the starting lengths of chamber was used, which is 23\,mm. Points closer to blue are nearer to the starting point, and in the theoretical model, an approximately spherical workspace with a radius of 6 mm was obtained. To ensure the device operated within a safe pressure range, we chose a spherical space with a radius not exceeding 5 mm as the experimental range.

\begin{figure}
    \centering
    \includegraphics[width=\linewidth]{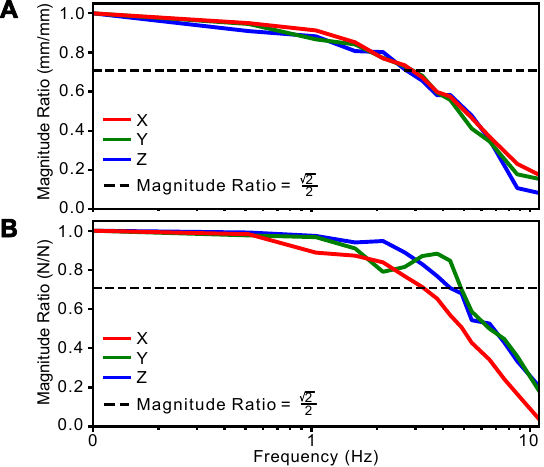}
    \caption{Bandwidth test for (A) movement and (B) block force.}
    \label{fig:bandwidth}
\end{figure}

\begin{figure*}[t]
    \vspace{1.7mm}
    \centering
    \includegraphics[width=\linewidth]{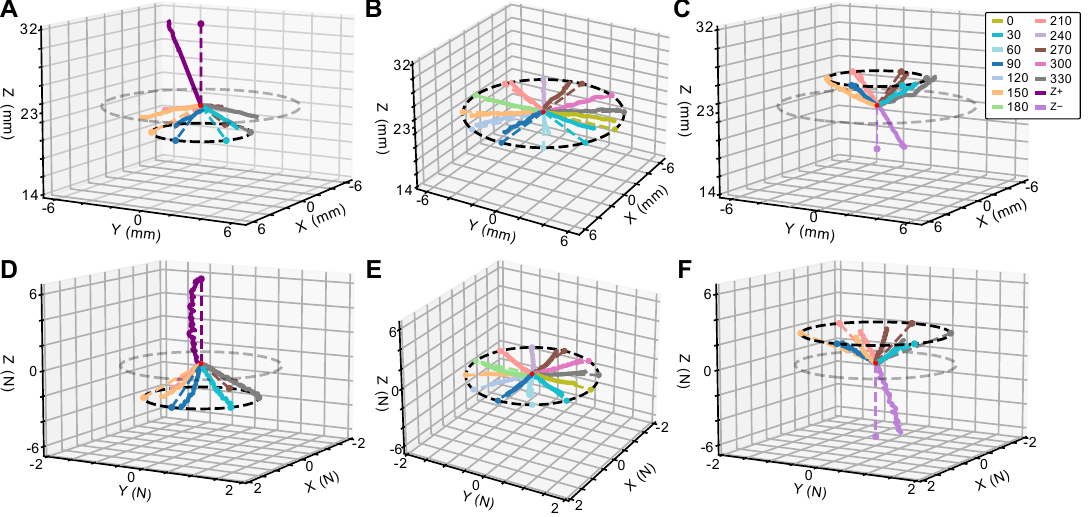}
    \caption{Commanded vs. actual position of the device control points at the (A) lower-, (B) middle-, and (C) upper-level paths. Commanded vs. actual block force direction up to actuator saturation for (D) lower-, (E) middle-, and (F) upper-level paths. Lower- and middle-level plots also indicate positive Z and negative Z paths. Middle-level plot views are elevated to improve visibility of the paths.}
    \label{fig:pos_force_output}
\end{figure*}


\changed{The radial range and path errors achieved over the commanded directions towards target points at the lower, middle and upper levels are shown in Tbl.\,\ref{tab:combined_ring}. In the free motion state, the minimum radial positional ranges of the \mbox{lower-,} \mbox{middle-}, and upper-level movements were 3.18\,mm, 4.48\,mm, and 3.88\,mm, respectively. In the vertical direction, our device achieved a minimum radial range of 4.89\,mm and 9.25\,mm in the Z$-$ and Z$+$ directions, respectively. In the blocked state, the minimum radial force range of the {lower-,} \mbox{middle-}, and upper-level movements were 1.94\,N, 1.01\,N, and 2.23\,N, respectively. The minimum force range was 6.01 and 6.79 in the Z$-$ and Z$+$ directions, respectively. Compared to the DC motor-actuated Foldaway origami haptic device  \cite{mintchev_portable_2019}, our soft haptic device has significant advantages in terms of dimensions and on-device weight, measuring 24\,mm $\times$ 24\,mm $\times$ 18\,mm and weighing 13\,g versus the Foldaway's dimensions and on-device weight of 90\,mm $\times$ 95\,mm $\times$ 22\,mm and 130\,g respectively. Our device exhibits a force range superior to Foldaway, achieving 6.01\,N in the vertical direction compared to 2\,N. \revised{However, the workspace range, position resolution, and bandwidth of our device are less than those of Foldaway, which are 35\,mm $\times$ 35\,mm $\times$ 30\,mm, 0.015mm and 20\,Hz, respectively. In comparison, our device has a workspace of approximately 10\,mm $\times$ 10\,mm $\times$ 10\,mm, with average positional error of 0.72mm, and a bandwidth of 3\,Hz. The lower bandwidth is typical of soft pneumatically actuated systems \cite{zhakypov_fingerprint_2022}, while the lower positional resolution arises from specific device characteristics which we further detail in the subsequent discussion.
}}

The results of the path following experiment are presented in Table\,\ref{tab:combined_ring}. The path error of the \mbox{lower-,} middle-, and upper-level movements averaged over all paths are 0.72\,mm, 0.63\,mm, and 0.47\,mm. For the negative and positive vertical directions the error is larger, at 1.02\,mm and 1.61\,mm, respectively. The just noticeable difference (JND) of the metacarpophalangeal joint of the human finger when the proximal interphalangeal joint is at a $45^\circ$ configuration is $2.7^\circ$\cite{tan_discrimination_2007}. The average lengths of the proximal, middle, and distal finger segments are 47.35\,mm, 25.37\,mm, and 23.81\,mm respectively \cite{Veber2006}. Thus the JND at the fingertip can be determined to be 1.74\,mm via the law of cosines and the arc length equation. Thus, the displacement error of the device tip in all directions is within the JND. Overall, our device displays better agreement with the piecewise constant curvature kinematic model for the middle and upper levels (Fig.\,\ref{fig:pos_force_output}B and C) compared to the lower level (Fig.\,\ref{fig:pos_force_output}C). The results of the force output path accuracy experiment are presented in \ref{tab:combined_ring}. The force output errors for the \mbox{lower-,} \mbox{middle-}, and upper-level force paths averaged over all paths are 0.15\,N, 0.14\,N, and 0.34\,N, respectively. In the negative and positive vertical directions, the force errors are 0.13\,N and 0.27\,N, respectively. For the lower and middle level, and positive vertical direction, the error is below the human finger force JND of 0.224\,N \cite{endo_force_2014}. However, the upper-level path error is 0.12N higher, while the positive vertical direction error is 0.046\,N higher than the JND.

There are several possible sources of errors in the device. \revised{When moving on paths for lower level that are close to its fully compressed configuration, the positional error with respect to the kinematic model could have occurred due to the self-interference caused by the air chambers and rigid components of the SPAs.} The force error was lower for the lower levels, because the blocked state constrained the device's movements and prevent the self-interference. On the contrary, the force error became more significant when reaching toward higher targets. This likely stemmed from the 3D-printed omnidirectional ball joint that connects the sensor to the device. Due to the printing accuracy and wear during the experiment, gaps appeared between the ball and socket, leading to uneven and loose contact. 

\hlight{Additionally, local stiffness variabilities, leading to radial asymmetry in the device's output force and movement shown in Fig.\,\ref{fig:pos_force_output}C, may have been introduced during the parallel SPA assembly process.} For example, uneven application of adhesive can lead to local stiffness variations in the device, causing different strain responses to the pneumatic load in each chamber, ultimately resulting in deviations in the magnitude and direction of the output force from the model predictions. The vertical position measurements were conducted last due to logistical issues, and the significantly higher error in the position output in these directions was due to the degradation of the actuators after repeated actuations, which led to distinct asymmetrical variation in the SPA pressure to length mappings. Besides the aforementioned errors, the precision and calibration errors of the pressure regulators also contributed to the deviation in vertical movement.

The device bandwidth was approximately 3\,Hz in free motion (Fig.\,\ref{fig:bandwidth}A) and under block force (Fig.\,\ref{fig:bandwidth}B). In the latter, the limiting bandwidth was observed to be in the x-direction, with the y- and z-directions having a bandwidth of 4\,Hz.
The bandwidth capabilities make the device suitable for haptic interactions where users move relatively slowly, or where the environment stiffness is low. Such interactions can be found in applications such as minimally invasive telesurgery where the required movements are careful and deliberate, and the user is interacting with soft deformable tissues. 

\changed{The performance of the device in free space compared to the computed interaction forces during interaction with a virtual cube is shown in Fig.\,\ref{fig:demo}C, the displacement of the tip recorded by the motion tracker closely follows the general trend of the force generated in CHAI3D. There is good consistency in both the x- and z-directions. However, a significant reduction in amplitude is observed in the y-direction. This is also due to the asymmetry introduced during the manufacturing and control processes, resulting in the stiffness of SPA being greater in the $y$-direction than in the $x$- and $z$-directions, as evidenced in Fig.\,\ref{fig:pos_force_output}B. Additionally, our device showcases good isolation of displacement in each of the Cartesian directions, with minimal residual displacements being commanded in unwanted directions during contact with each side of the cube.
}



In our current device, explicit static force control using the Jacobian was not performed. This was because the mapping from pressure to axial force change is difficult to analytically model for the complex SPA geometry and material. In the future we plan to investigate methods to enable explicit force control, including using finite element analysis to determine the pressure to axial force mapping of the SPA, dynamic modeling of the soft actuator \cite{falkenhahn_dynamic_2015}, and embedding soft force and position sensors for closed-loop control \cite{mccandless_soft_2023,choi2023integrated}. Introducing position sensing in particular will further enable the device to function as a human interface, instead of purely as a haptic display.

Additionally, we will investigate augmenting the device to enable more haptic functionality, such as passive stiffness tuning using jamming mechanisms \cite{giri_hapstick_2023}. Given the high power density of the pneumatic actuation, we will further aim to develop handheld interfaces with a device for each finger to enable multi-digit feedback for tasks such as grasping \cite{feix_grasp_2016}.


\section{CONCLUSION}
In this paper, we proposed a novel soft pneumatic handheld haptic device that can provide three degree-of-freedom kinesthetic force feedback to the index finger. In particular, we envision that device can be used for delivery sensory substitution to mitigate instabilities inherent in direct kinesthetic force feedback. We modeled the device using the constant curvature assumption and used it to perform open-loop model-based control. We validated the output range, accuracy, and bandwidth of the haptic device in both free motion and blocked states, achieving positional and force precision below or close to the human JND. The high power density of the pneumatic systems allows for high force output in a novel compact and lightweight form factor,  showcasing its promising future application in grounded and handheld human-machine interfaces that deliver single- or multi-digit interaction and feedback.


\addtolength{\textheight}{-0cm}   



\section*{ACKNOWLEDGMENT}
The authors thank Jianfeng Zhou for his help with the mechatronic control hardware and figure-making, and Sai Jiang and Juan Beaver for their help with the CHAI3D simulation.

\bibliographystyle{IEEEtran}
\bibliography{bibliography-main,zong-references}










\end{document}